\documentclass[twoside]{article}

\usepackage[accepted]{aistats2020}

\usepackage[round]{natbib}

\usepackage{url}
\usepackage{defs}
\usepackage{amsthm}
\usepackage{xcolor}

\newcommand\bbetahat{\widehat{\bbeta}}
\newcommand\ens{\mathrm{ens}}
\newcommand\ridge{\mathrm{ridge}}
\newcommand\bias{\mathrm{bias}}
\newcommand\variance{\mathrm{variance}}
\newcommand\bbetaprime{\bbeta^\prime}
\newcommand\setS{\mathcal{S}}
\newcommand\setT{\mathcal{T}}

\newcommand\transp{\top}

\newtheorem{theorem}{Theorem}[section]
\newtheorem{lemma}[theorem]{Lemma}
\newtheorem{proposition}[theorem]{Proposition}
\newtheorem{corollary}[theorem]{Corollary}

\newcounter{preview}
\newtheorem{likeridgepreview}{Theorem}[preview]

\theoremstyle{definition}
\newtheorem{assumption}[theorem]{Assumption}

\usepackage{balance}

\begin{document}

\twocolumn[

\aistatstitle{The Implicit Regularization of Ordinary Least Squares Ensembles}

\aistatsauthor{ Daniel LeJeune \And Hamid Javadi \And  Richard G. Baraniuk }
\aistatsaddress{ Rice University \And  Rice University \And Rice University } ]

\begin{abstract}
    Ensemble methods that average over a collection of independent
    predictors that are each limited to a subsampling of both the examples
    and features of the training data command a significant presence in
    machine learning, such as the ever-popular random forest, yet the
    nature of the subsampling effect, particularly of the features, is not
    well understood. We study the case of an {\em ensemble of linear
    predictors}, where each individual predictor is fit using ordinary
    least squares on a random submatrix of the data matrix. We show that,
    under standard Gaussianity assumptions, when the number of features
    selected for each predictor is optimally tuned, the asymptotic risk of
    a large ensemble is equal to the asymptotic {\em ridge regression}
    risk, which is known to be optimal among linear predictors in this
    setting. In addition to eliciting this {\em implicit regularization}
    that results from subsampling, we also connect this ensemble to the
    dropout technique used in training deep (neural) networks, another
    strategy that has been shown to have a ridge-like regularizing effect.
\end{abstract}

\section{INTRODUCTION}

\emph{Ensemble methods} \citep{Breiman1996, amit1997shape, Josse:2016:BRL:2946645.3007077} are an oft-used strategy employed successfully in a broad range of problems in machine learning and statistics, in which one combines a number of \emph{weak predictors} together to obtain one powerful predictor. This is accomplished by giving each weak learner a different \emph{view} of the training data. Various strategies for changing this training data view exist, among which many are simple sampling-based techniques in which each predictor is (independently) given access to a subsampling of the rows (examples) and columns (features) of the training data matrix, such as \emph{bagging} \citep{Breiman1996, buhlmann2002}. Another noteworthy technique is \emph{boosting} \citep{FREUND1997119, breiman1998}, in which the training data examples are reweighted adaptively according to how badly they have been misclassified while building the ensemble. 
In this work, we consider the former class of techniques---those that train each weak predictor using an independent subsampling of the training data.

\begin{figure}[t]
    \centering
    \includegraphics{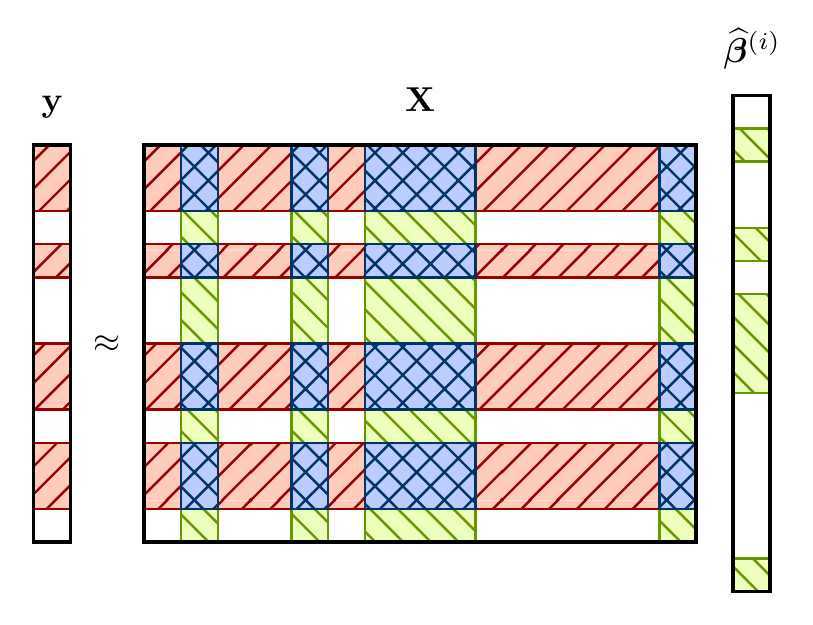}
    \caption{
    Example (rows) and feature (columns) subsampling of the training data used in the ordinary least squares fit for one member of the ensemble. The $i$-th member of the ensemble is only allowed to predict using its subset of the features (green). It must learn its parameters $\bbetahat{}^{(i)}$ by performing ordinary least squares using the subsampled examples of $\vy$ (red) and the subsampled examples (rows) and features (columns) of the data matrix $\mX$ (blue, crosshatched).
}
    \label{fig:subsampling}
\end{figure}

Ensemble methods based on independent example and feature subsampling are attractive for two reasons. First, they are computationally appealing in that they are massively parallelizable, and since each member of the ensemble uses only part of the data, they are able to overcome memory limitations faced by other methods \citep{10.1007/978-3-642-33460-3_28}. 
Second, ensemble methods are known to achieve lower risk due to the fact that combining several different predictors reduces variance \citep{buhlmann2002,JMLR:v15:wager14a, scornet2015}, and empirically they have been found to perform very well. \emph{Random forests} \citep{Breiman2001, athey2019, friedberg2018local}, for example, ensemble methods that combine example and feature subsampling with decision trees by choosing the most useful feature from a random subset of the features at each branch of the tree, remain among the best-performing off-the-shelf machine learning methods available \citep{Cutler01pert-, JMLR:v15:delgado14a, JMLR:v18:15-240}.

Let $\mX \in \reals^{n \times p}$ be the training data matrix consisting of $n$ examples of data points each having $p$ features.
While there exist theoretical results on the benefits of \emph{example (row) subsampling} \citep{buhlmann2002}, 
the exact nature of the effect of \emph{feature (column) subsampling} on ensemble performance remains poorly understood.
In this paper, we study the prototypical form of this problem in the context of linear regression. That is, given the data matrix $\mX$ and target variables $\vy \in \reals^n$, we study the ensemble $\bbetahat{}^\ens = \frac{1}{k} \sum_{i=1}^k \bbetahat{}^{(i)}$, where each $\bbetahat{}^{(i)}$ is learned using ordinary least squares on an independent random subsampling of both the examples and features of the training data. This subsampling is illustrated in Figure~\ref{fig:subsampling}. 
We show that under such a scheme, the resulting predictor of this ensemble performs as well as the \emph{ridge regression} \citep{Hoerl1970ridge, friedman2001elements} predictor fit using the entire training data, which is known to be the optimal linear predictor under the data assumptions that we consider. 
Further, the asymptotic risk of the ensemble depends \emph{only on the amount of feature subsampling} and not on the amount of example subsampling, provided each individual ordinary least squares problem is underdetermined.
Our main result in Theorem~\ref{thm:like-ridge}
can be summarized as follows:

\begin{likeridgepreview}[informal statement]
    When the features and underlying model weights both follow i.i.d.\ Gaussian distributions, the optimal asymptotic risk for an ensemble of ordinary least squares predictors is equal to the optimal asymptotic ridge regression risk. 
\end{likeridgepreview}

We can interpret this result as an example of \emph{implicit regularization} \citep{mahoney2012pods,neyshabur2014search,Gunasekar2017implicit,arora2019implicit}. That is, while the individual ordinary least squares subproblems are completely unregularized, the ensemble behaves as if it had been regularized using a ridge regression penalty. Recently, there has been much interest in investigating the implicit regularization effects of commonly used heuristic methods, particularly in cases where they enable the training of highly \emph{overparameterized} models that generalize well to test data despite having the capacity to overfit the training data \citep{Zhang2017understanding, belkin2018interpolate}. Examples of heuristic techniques that have been shown to have implicit regularization effects include stochastic gradient descent \citep{pmlr-v48-hardt16} and \emph{dropout} \citep{Srivastava2014dropout}. Incidentally, we show a strong connection between the ensemble of ordinary least squares predictors and dropout, which is known to have a ridge-like regularizing effect \citep{wager2013dropout}, and we make this link via stochastic gradient descent.

\paragraph{Contributions}
We summarize our contributions as follows:
\textbf{[C1]} We prove that when the amount of feature subsampling is optimized to minimize risk, an ensemble of ordinary least squares predictors achieves the same risk as the optimal ridge regression predictor asymptotically as $n, p \to \infty$ (see Section~\ref{sec:ensemble-risk}).
\textbf{[C2]}~We demonstrate the converge of the ensemble risk to the optimal ridge regression risk via simulation (see Section~\ref{sec:convergence}). 
\textbf{[C3]}~We reveal a connection between the ordinary least squares ensemble and the popular \emph{dropout} technique used in deep (neural) network training (see Section~\ref{sec:dropout}) and from the insight gained from this connection develop a recipe for mitigating excess risk under suboptimal feature subsampling via simple output scaling (see Section~\ref{sec:scaled-ensemble}).

\section{ENSEMBLES OF ORDINARY LEAST SQUARES PREDICTORS}

We consider the familiar setting of \emph{linear regression}, where there exists a linear relationship between the target variable $y \in \reals$ and the feature variables $\vx \in \reals^p$---i.e., $y = \langle \vx, \bbeta \rangle$, where $\bbeta \in \reals^p$ is the model parameter vector. The goal of a machine learning algorithm is to estimate these parameters given $n$ i.i.d.\ noisy samples $\set{\vx^{(i)}, y^{(i)}}_{i=1}^n$. The noise relationship is given by
\begin{align}
    \vy = \mX \bbeta + \sigma \vz,
\end{align}
where $[\mX]_{ij} = [\vx^{(i)}]_j$, $[\vy]_i = y^{(i)}$, and $[\vz]_i = z^{(i)}$, where $z^{(i)}$ are i.i.d.\ zero-mean random variables with unit variance independent of $\mX$. We assume a Gaussian $\mathcal{N}({\bm 0}, \bSigma)$ distribution on $\vx$, and for the results in this paper, we assume $\bSigma = \mI_p$.

Our ensemble consists of $k$ linear predictors each fit using ordinary least squares on a submatrix of $\mX$, and the resulting prediction is the average of the outputs. Equivalently, our ensemble is defined by its estimate of the parameters
\begin{align}
    \bbetahat{}^\ens \defeq \frac{1}{k} \sum_{i=1}^k \bbetahat{}^{(i)},
\end{align}
where $\bbetahat{}^{(i)}$ is the parameter estimate of the $i$-th member of the ensemble. To characterize the estimates $\bbetahat{}^{(i)}$, we first introduce some notation. 
Let the \emph{selection matrix} $\mS$ corresponding to a subset of indices $S \subseteq [p]$, where $[p] = \set{1, \ldots, p}$, denote the the $p \times |S|$ matrix obtained by selecting from $\mI_p$ the columns corresponding to the indices in $S$, where $\mI_p$ denotes the $p \times p$ identity matrix. 
With this definition of selection matrices, for $S \subseteq [p]$ and $T \subseteq [n]$, we have that $\mT^\transp \mX \mS$ is the matrix of size $|T| \times |S|$ obtained from $\mX$ by selecting (subsampling) the rows and columns indicated by sets $T$ and $S$.
Returning to the ensemble, let $\setS \defeq (S_i)_{i=1}^k$ and $\setT \defeq (T_i)_{i=1}^k$ denote the collection of \emph{feature subsets} and \emph{example subsets}, respectively, where each $S_i \subseteq [p]$ and each $T_i \subseteq [n]$. Then, assuming $|S_i| < |T_i|$, for each member of the ensemble we let
\begin{align}
    \bbetahat{}^{(i)}_{S_i} &= \argmin_{\bbetaprime{}} \norm[2]{\mT_i^\transp \left(\mX \mS_i \bbetaprime - \vy \right)}, \\
    \bbetahat{}^{(i)}_{S_i^c} &= {\bm 0},
\end{align}
where $S_i^c = [p] \setminus S_i$ denotes the complement of the set $S_i$.
This can alternatively be written in closed form as
\begin{align}
    \label{eq:pseudoinverse-solution}
    \bbetahat{}^{(i)} = \mS_i \left(\mT_i^\transp \mX \mS_i\right)^\dagger \mT_i^\transp \vy,
\end{align}
where $(\cdot)^\dagger$ denotes the Moore--Penrose pseudoinverse. Thus, the closed-form expression for the ensemble parameter estimate is given by
\begin{align}
    \bbetahat{}^\ens = \frac{1}{k} \sum_{i=1}^k \mS_i \left(\mT_i^\transp \mX \mS_i\right)^\dagger \mT_i^\transp \vy.
\end{align}

\section{ENSEMBLE RISK}
\label{sec:ensemble-risk}

We define the \emph{risk} of a linear predictor as the expected squared error of a prediction of the target variable on an independent data point $\vx$:
\begin{align}
    R(\bbetaprime) &\defeq 
    \E_\vx \left[ 
    \left\langle
    \vx, \bbeta - \bbetaprime
    \right\rangle^2
    \right] \nonumber \\
    &= \left\langle
    \bbeta - \bbetaprime, 
    \bSigma \left(\bbeta - \bbetaprime \right)
    \right\rangle.
\end{align}
For any predictor of the form $\bbetaprime{} = f(\mX) \vy$, for some $f : \reals^{n \times p} \to \reals^{p \times n}$, we can rewrite parameter estimation error as
\begin{align}
    \bbeta - \bbetaprime{} = 
    (\mI_p - f(\mX)\mX) \bbeta - \sigma f(\mX)\vz.
\end{align}
Then by the independence of $\mX$ and $\vz$ and some algebra, we can decompose the risk into the so-called ``bias'' and ``variance'' components
\begin{align}
    \label{eq:expected-risk}
    \E_\vz \left[ R(\bbetaprime{}) \right]
    ={} &\underbrace{\left \langle
    \bbeta \bbeta^\transp, 
    (\mI_p - f(\mX)\mX)^\transp \bSigma (\mI_p - f(\mX)\mX)
    \right \rangle}_{\bias(\bbetaprime{})} \nonumber \\
    &+ \underbrace{\sigma^2 \left\langle 
    f(\mX), \bSigma f(\mX)
    \right\rangle}_{\variance(\bbetaprime{})}.
\end{align}
For the ensemble, we obtain for the bias and variance
\begin{align}
    \label{eq:bias-sum}
    \bias(\bbetahat{}^\ens) 
    &= \frac{1}{k^2} 
    \sum_{i, j=1}^k 
    \bias_{ij}(\bbetahat{}^\ens) \\
    \label{eq:variance-sum}
    \variance(\bbetahat{}^\ens)
    &= \frac{1}{k^2} 
    \sum_{i, j=1}^k 
    \variance_{ij}(\bbetahat{}^\ens),
\end{align}
where
\begin{align}
    \bias_{ij}(\bbetahat{}^\ens)
    &={}  \Big\langle
    \bbeta \bbeta^\transp, 
    \left(
    \mI_p - \mS_i \left(\mT_i^\transp \mX \mS_i\right)^\dagger \mT_i^\transp \mX 
    \right)^\transp \nonumber \\
    & \times
    \bSigma
    \left(
    \mI_p - \mS_j \left(\mT_j^\transp \mX \mS_j\right)^\dagger \mT_j^\transp \mX 
    \right)
    \Big\rangle,
\end{align} 
\vspace{-2em}
\begin{align}
    \label{eq:variance-ij}
    \variance_{ij}(\bbetahat{}^\ens) 
    ={}  \sigma^2 \Big\langle
    &\mS_i \left(\mT_i^\transp \mX \mS_i\right)^\dagger \mT_i^\transp,
    \nonumber \\
    &\bSigma \mS_j \left(\mT_j^\transp \mX \mS_j\right)^\dagger \mT_j^\transp
    \Big\rangle.
\end{align}
Thus, evaluating the risk of the ensemble is a matter of evaluating these pairwise interaction terms.

To begin evaluating the above terms, we need to introduce additional assumptions. Specifically, we assume that the subsets are independent and that all indices are equally likely to be included in each subset. 

\begin{assumption}[finite subsampling]
    \label{assumption:finite-subsampling}
    The subsets in the collections $\setS$ and $\setT$ are selected at random such that  $|S_i| < |T_i| - 1$ and that the following hold:
    \begin{itemize}
        \item $\Pr(j \in S_i) = |S_i| / p$ for all $j \in [p]$,
        \item $\Pr(m \in T_i) = |T_i| / n$ for all $m \in [n]$,
        \item The subsets $S_1, S_2, \ldots, S_k, T_1, T_2, \ldots, T_k$ are conditionally independent given the example subset sizes $(|T_i|)_{i=1}^k$.
    \end{itemize}
\end{assumption}
A simple sampling strategy that satisfies these assumptions is to fix $|S_i|$ and $|T_i|$ such that $|S_i| < |T_i| - 1$ and select subsets uniformly at random of the given sizes. Another strategy is to construct the subsets by flipping a coin for each index, rejecting any resulting subsets that fail to satisfy $|S_i| < |T_i| - 1$.

With Assumption~\ref{assumption:finite-subsampling}, we are now equipped to evaluate the pairwise interaction terms. The following two lemmas enable us to characterize the bias and variance components of the risk in the finite-dimensional setting. The proofs of these lemmas are exercises in linear algebra and conditional expectations and can be found in the Appendix.
 
With some slight abuse of notation, we allow $\E_{\setS, \setT}$ to denote the expectation taken with respect to the choice of indices in the subsets, but not their sizes.
In other words, $\E_{\setS, \setT}$ indicates the conditional expectation over $\setS$ and $\setT$, conditioned on the subset sizes indicated by the context.

\begin{lemma}[bias]
    \label{lemma:bias}
    Assume that $\bSigma = \mI_p$ and that Assumption~\ref{assumption:finite-subsampling} holds. Then
\begin{multline}
    \label{eq:bias-finite}
    \E_{\mX, \setS, \setT} \left[ \bias_{ij}(\bbetahat{}^\ens) \right]
    \\
    = \begin{cases}
    \frac{|S_i^c \cap S_j^c|}{p}
    \left(1 + \frac{|S_i \cap S_j|}{n - |S_i \cap S_j| - 1} 
    \right)
    \norm[2]{\bbeta}^2 & {\rm{if }}\; i \neq j, \\
    \frac{|S_i^c|}{p}
    \left(1 + \frac{|S_i|}{|T_i| - |S_i| - 1} 
    \right)
    \norm[2]{\bbeta}^2 & {\rm{if }}\; i = j.
    \end{cases}
\end{multline}
\end{lemma}

\begin{lemma}[variance]
    \label{lemma:variance}
    Assume that $\bSigma = \mI_p$ and that Assumption~\ref{assumption:finite-subsampling} holds. Then
\begin{align}
    \label{eq:variance-finite}
    &\E_{\mX, \setS, \setT} \left[ \variance_{ij}(\bbetahat{}^\ens) \right]
    = \begin{cases}
    \frac{\sigma^2 |S_i \cap S_j|}{n - |S_i \cap S_j| - 1} & {\rm{if }}\; i \neq j, \\
    \frac{\sigma^2 |S_i|}{|T_i| - |S_i| - 1} & {\rm{if }}\; i = j.
    \end{cases}
\end{align}
\end{lemma}
One observation that we can make already from these results is that the example subsampling only affects the terms where $i=j$. Assuming that the subsampling procedure is the same for each $i$, so that for large $k$ the $i \neq j$ terms are sure to dominate the sum, this means that in the limit as $k \to \infty$, the effects of example subsampling are non-existent. We note that this is a result of the assumption that $|S_i| < |T_i|$, and that if we were to have $|S_i| > |T_i|$, then we would observe effects of example subsampling when $i \neq j$, which we discuss further in Section~\ref{sec:fat-submatrices}.

We now turn our attention to the setting where $n, p \to \infty$ in order to better reason about the results contained in these lemmas. We introduce the following additional assumption.
\begin{assumption}[asymptotic subsampling]
    \label{assumption:asymptotic-subsampling}
    For some $\alpha, \eta \in [0, 1]$, the subsets in the collections $\setS$ and $\setT$ are selected randomly such that $|S_i|/p \xrightarrow{{\rm{a.s.}}} \alpha$ as $p \to \infty$ and $|T_i|/n \xrightarrow{{\rm{a.s.}}} \eta$ as $n \to \infty$ for all $i \in [k]$.
\end{assumption}
This assumption is easily satisfied. For example, in the sampling strategy where we fix $|S_i|$ and $|T_i|$, we can choose $|S_i| = \floor{\alpha p}$ and $|T_i| = \floor{\eta n}$. For the coin-flipping strategy, we can select feature subsets with a coin of probability $\alpha$ and example subsets with a coin of probability $\eta$.

Under this assumption, and additionally assuming without loss of generality that $\norm[2]{\bbeta} = 1$, if $n, p \to \infty$ such that $p/n \to \gamma$ and $\eta > \alpha\gamma$, the quantities in \eqref{eq:bias-finite} and \eqref{eq:variance-finite} converge almost surely as follows:
\begin{multline}
    \label{eq:bias-limiting}
    \E_{\mX, \setS, \setT} \left[ \bias_{ij}(\bbetahat{}^\ens) \right]
    \\
    \xrightarrow{a.s.} \begin{cases}
    (1 - \alpha)^2
    \left(1 + \frac{\alpha^2 \gamma}{1 - \alpha^2 \gamma} 
    \right)
    & \text{if } i \neq j, \\
    (1 - \alpha)
    \left(1 + \frac{\alpha \gamma}{\eta - \alpha \gamma} 
    \right)
    & \text{if } i = j,
    \end{cases}
\end{multline}
and
\begin{align}
    \label{eq:variance-limiting}
    &\E_{\mX, \setS, \setT} \left[ \variance_{ij}(\bbetahat{}^\ens) \right]
    \xrightarrow{a.s.} \begin{cases}
    \frac{\sigma^2 \alpha^2 \gamma}{1 - \alpha^2 \gamma} & \text{if } i \neq j, \\
    \frac{\sigma^2 \alpha \gamma}{\eta - \alpha \gamma} & \text{if } i = j.
    \end{cases}
\end{align}
We are now equipped to state our asymptotic risk result for the ensemble of ordinary least squares predictors. Denote for an ensemble satisfying Assumptions~\ref{assumption:finite-subsampling} and \ref{assumption:asymptotic-subsampling} with parameters $\alpha$, $\eta$, and $k$ the \emph{limiting risk}
\begin{align}
    R_{\alpha, \eta, k}^\ens \defeq \lim_{n, p \to \infty} \E_{\mX, \vz, \setS, \setT} \left[ R(\bbetahat{}^\ens) \right].
\end{align}
From \eqref{eq:bias-sum} and \eqref{eq:variance-sum}, we know that both the bias and variance components of the limiting risk are the averages of $k^2$ terms, and from \eqref{eq:bias-limiting} and \eqref{eq:variance-limiting}, we know that the $k(k-1)$ terms where $i \neq j$ will take one value and the remaining $k$ terms where $i=j$ will take another. Thus we have the \emph{limiting bias}
\begin{multline}
    \label{eq:limiting-bias}
    \lim_{n, p \to \infty} \E_{\mX, \setS, \setT} \left[ \bias(\bbetahat{}^\ens) \right]
    \\
    = \frac{k-1}{k} \left( \frac{(1 - \alpha)^2}{1 - \alpha^2 \gamma }\right)
    + \frac{1}{k} \left( \frac{\eta(1 - \alpha)}{\eta - \alpha \gamma }\right)
\end{multline}
and \emph{limiting variance}
\begin{multline}
    \label{eq:limiting-variance}
    \lim_{n, p \to \infty} \E_{\mX, \setS, \setT} \left[ \variance(\bbetahat{}^\ens) \right] \\
    = \frac{k-1}{k} \left( \frac{\sigma^2 \alpha^2 \gamma}{1 - \alpha^2 \gamma }\right)
    + \frac{1}{k} \left( \frac{\sigma^2 \alpha \gamma}{\eta - \alpha \gamma }\right).
\end{multline}
Upon careful examination of these quantities, we observe that in fact both the limiting bias \emph{and} the limiting variance are \emph{decreasing} in $k$, and thus the ensemble serves not only as a means to reduce variance (as is well understood), \emph{but also} to reduce bias. We defer further discussion to Section~\ref{sec:bias-variance}.
Adding the limiting bias and variance together yields the following result.
\begin{theorem}[limiting risk]
    \label{thm:ensemble-risk}
    Assume that $\bSigma = \mI_p$ and ${\norm[2]{\bbeta} = 1}$ and that Assumptions \ref{assumption:finite-subsampling} and \ref{assumption:asymptotic-subsampling} hold. Then in the limit as $n, p \to \infty$ with $p/n \to \gamma$, for $\eta > \alpha \gamma$, we have almost surely that
\begin{multline}
    R_{\alpha, \eta, k}^\ens
    = \frac{k-1}{k} 
    \left(
    \frac{(1 - \alpha)^2 + \sigma^2 \alpha^2 \gamma}{1 - \alpha^2 \gamma} 
    \right) 
    \\
    + \frac{1}{k}
    \left(
    \frac{\eta(1-\alpha) + \sigma^2 \alpha \gamma}{\eta - \alpha \gamma}
    \right).
\end{multline}
\end{theorem}
Here we see again more explicitly that for large $k$, the effect of example subsampling vanishes. This leaves us with the \emph{large-ensemble} risk
\begin{align}
\label{eq:large-ensemble}
    R_{\alpha}^\ens{} 
    &\defeq \lim_{k \to \infty} R_{\alpha, \eta, k}^\ens 
    \nonumber \\
    &= \frac{(1 - \alpha)^2 + \sigma^2 \alpha^2 \gamma}{1 - \alpha^2 \gamma}.
\end{align}
We note that while the large-ensemble risk depends only upon $\alpha$, we cannot realize this risk with an ensemble if $\eta \leq \alpha \gamma$. Our remaining results concern the large-ensemble risk and therefore assume that $\eta = 1$ for simplicity, but we caution the reader that some of these results may not be valid for some smaller values of $\eta$, depending on $\sigma$ and $\gamma$.

Because $\alpha$ is an algorithmic hyperparameter, it can be tuned to minimize the risk. If we do so, then what we obtain is the perhaps surprising result that the optimal large-ensemble risk of the ordinary least squares predictor is equal to the limiting risk of the \emph{ridge regression} predictor under our assumptions. The ridge regression predictor with parameter $\lambda$ is defined as
\begin{align}
    \bbetahat{}_\lambda^\ridge{} 
    &\defeq \argmin_{\bbetaprime{}} \norm[2]{\mX \bbetaprime{} - \vy}^2 + \lambda \norm[2]{\bbetaprime{}}^2
    \nonumber \\
    &= \inv{\mX^\transp \mX + \lambda \mI_p} \mX^\transp \vy.
\end{align}
We formally state this result in the following theorem, which leverages the recent analysis of the limiting risk of ridge regression by \citet{dobriban2018}.\footnote{We note that results on MMSE estimation error from the wireless communication community \citep[see, e.g.,][]{tulino2004random} predate the more general result of \citet{dobriban2018}, and that these apply to the $\bSigma = \mI_p$ setting we consider, where risk is equal to estimation error.}
The proof is found in the Appendix.
\begin{theorem}
    \label{thm:like-ridge}
    Assume that $\bSigma = \mI_p$ and $\bbeta \sim \mathcal{N}({\bm 0}, p^{-1} \mI_p)$ and that Assumptions~\ref{assumption:finite-subsampling} and \ref{assumption:asymptotic-subsampling} hold with $\eta = 1$. Then in the limit as $n, p \to \infty$ with $p/n \to \gamma$, we have almost surely that
\begin{align}
    \inf_{\alpha < \gamma^{-1}} 
    R_{\alpha}^\ens
    = \inf_\lambda R \left(\bbetahat^{\ridge}_{\lambda} \right).
\end{align}
\end{theorem}

The implication of Theorem \ref{thm:like-ridge} is quite strong. Under the assumption of the theorem that true parameters $\bbeta$ have a Gaussian distribution with covariance $p^{-1}\mI_p$, the ridge regression predictor (the maximum a posteriori estimator for this setting) is the predictor with the lowest expected risk of all predictors of the form $\bbetaprime{} = f(\mX) \vy$. To see this, note that if we take the expectation of \eqref{eq:expected-risk} with respect to $\bbeta$, we find that the optimal $f(\mX)$ must satisfy the first order optimality condition
\begin{align}
    \bSigma f(\mX) (\mX \mX^\transp + p \sigma^2 \mI_p) = \bSigma \mX^\transp,
\end{align}
which for invertible $\bSigma$ yields the optimally tuned ridge regression predictor. Thus, in the $\bSigma = \mI_p$ setting, the optimally tuned ensemble achieves the optimal risk for any linear predictor.

A curious result obtained during the proof of this theorem is the following corollary relating the optimal large ensemble risk to the optimal choice of the hyperparameter $\alpha$.
\begin{corollary}
    \label{corollary:one-minus-alpha}
    Assume that $\bSigma = \mI_p$ and ${\norm[2]{\bbeta} = 1}$ and that Assumptions \ref{assumption:finite-subsampling} and \ref{assumption:asymptotic-subsampling} hold with $\eta = 1$. Then in the limit as $n, p \to \infty$ with $p/n \to \gamma$, we have almost surely that
    \begin{align}
        R_{\alpha_*}^\ens
        = 1 - \alpha_*,
    \end{align}
    where $\alpha_* = \argmin_{\alpha < \gamma^{-1}} R_\alpha^\ens$.
\end{corollary}
\section{DISCUSSION}

\subsection{Convergence}
\label{sec:convergence}

\begin{figure}[t]
    \centering
    \includegraphics{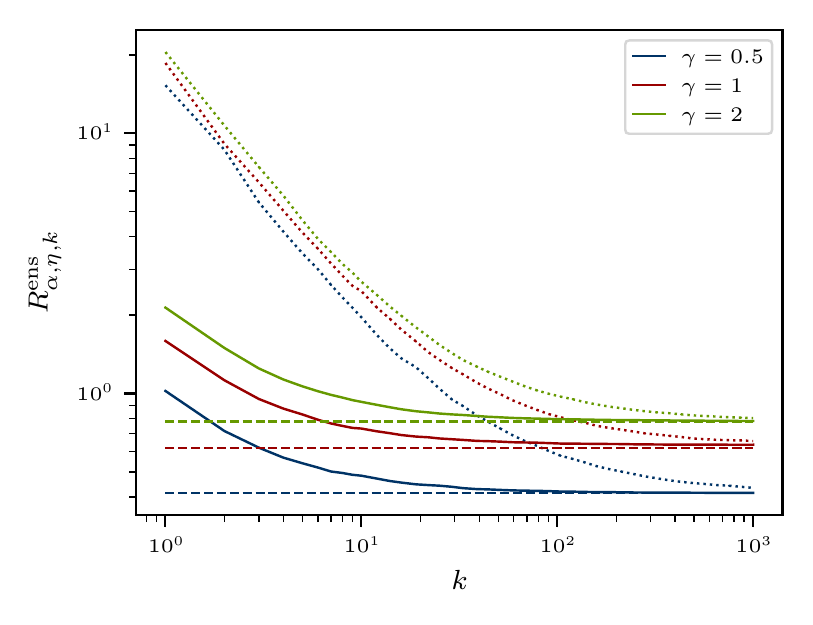}
    \caption{Approximate limiting risk (averaged over 50 trials with $n=200, \sigma=1$) when using $\eta = 1$ (solid) and $\eta = 1.1 \times \alpha \gamma$ (dotted). For each value of $\gamma$, both ensembles converge to the theoretical optimal ridge regression risk (dashed).}
    \label{fig:risk-versus-k}
\end{figure}

\begin{figure}[t]
    \centering
    \includegraphics{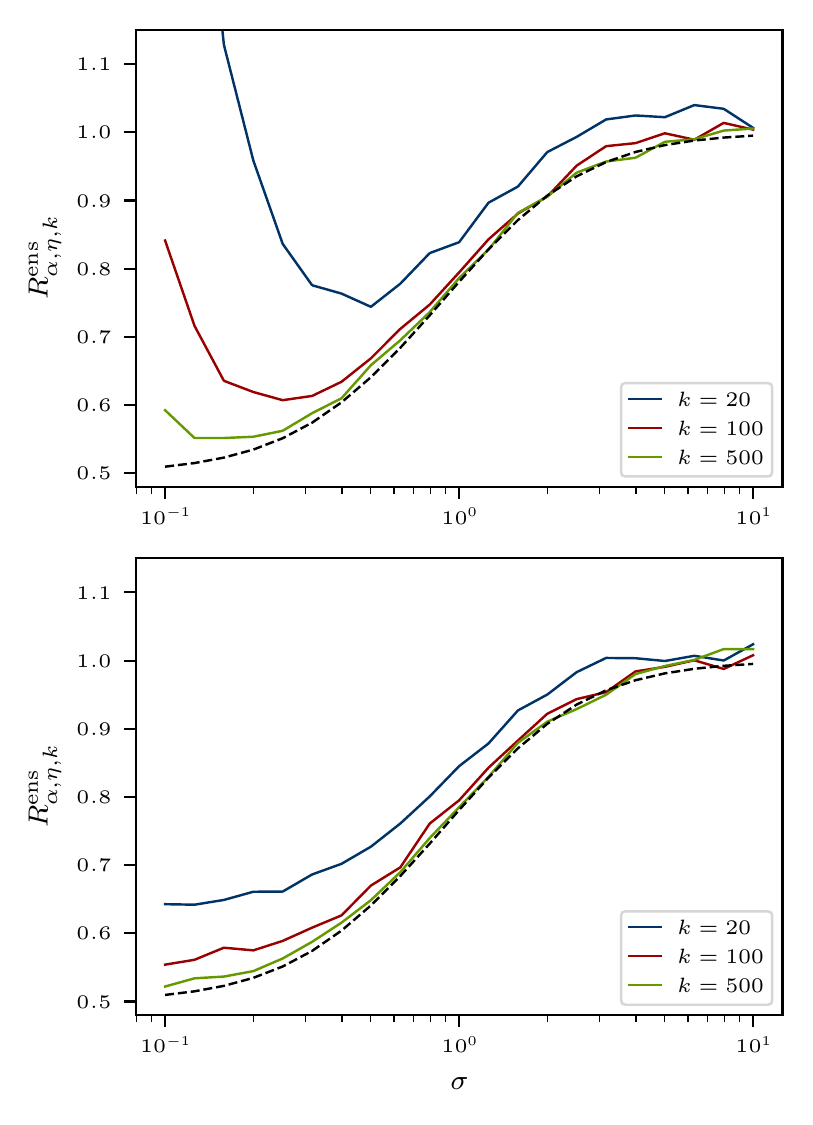}
    \caption{Approximate limiting risk (averaged over 100 trials with $n=200, p=400$) when using $\alpha=\alpha_*$ (top) and $\alpha = \argmin_{\alpha'} R_{\alpha', \eta, k}^\ens$ (bottom). As $k$ increases, in both cases the risk converges to the theoretical optimal ridge regression risk (black dashed).}
    \label{fig:risk-versus-sigma}
\end{figure}

In practice, any ensemble will have only a finite number of members. Therefore, it is important to understand the rates at which the risk of the ensemble converges to large-ensemble risk in \eqref{eq:large-ensemble}.
From Theorem~\ref{thm:ensemble-risk}, it is clear that as a function of $k$, the limiting risk converges to the large-ensemble risk at a rate $O(1/k)$. However, as the choice of $\eta$ approaches $\alpha \gamma$, this rate becomes slower. In Figure~\ref{fig:risk-versus-k}, we plot\footnote{See \url{https://github.com/dlej/ensemble-ols}.} the convergence in $k$ of the limiting risk to the large-ensemble risk for $\eta = 1$ (using all examples) and for $\eta = 1.1 \times (\alpha \gamma)$ (near to as small as possible while still having $|S_i| < |T_i|$). We plot these curves for $\sigma = 1$ and for three different values of $\gamma$, using $n=200$, which is sufficient to realize the convergence in $n$ and $p$. We choose $\alpha=\alpha_*$, the minimizer of the large-ensemble risk. What we observe is that, indeed, for both choices of $\eta$, the risks converge to the optimal ridge risk. As expected, however, with the smaller choice of $\eta$ the risk converges nearly an order of magnitude more slowly.

While the choice of $\alpha=\alpha_*$ will result in optimal risk for large enough ensembles, for finite $k$ this choice can in some cases be undesirable. For instance, consider the setting where $\eta = 1$ and $\gamma > 1$. Then as $\sigma \to 0$, $\alpha_* \to \gamma^{-1}$ (see expressions for $\alpha_*$ in the Appendix). This obviously yields the optimal large-ensemble risk, by definition, but for any finite $k$, the limiting risk tends to infinity for this choice of $\alpha$. However, if we know what the size of our ensemble will be, we can tune $\alpha$ to the limiting risk for finite $k$ instead of the large ensemble risk. In general, this means choosing an $\alpha$ smaller than $\alpha_*$.
In Figure~\ref{fig:risk-versus-sigma}, we demonstrate the convergence in $k$ to the large-ensemble risk as a function of $\sigma$ for $\alpha = \alpha_*$ and for $\alpha = \argmin_{\alpha'} R_{\alpha', \eta, k}^\ens$. We plot these curves for $\gamma = 2$ and $\sigma \in [0.1, 10]$, using $n = 200$. While for both choices of $\alpha$ we see convergence in $k$ for each $\sigma$, as $\sigma \to 0$, the risk is very large for $\alpha = \alpha_*$. For $\alpha$ adapted to the choice of $k$, however, this effect is mitigated.

\subsection{Bias and Variance Decrease with Ensemble Size}
\label{sec:bias-variance}

We return here to the observation made in Section~\ref{sec:ensemble-risk} that the limiting bias and variance are both \emph{decreasing} in $k$. 
Thus, although there is a bias--variance tradeoff in $\alpha$, there is no such tradeoff with $k$.
This can be seen by comparing the $i=j$ and $i \neq j$ terms in each case. In the case of bias, for the bias to be decreasing, it must be that
\begin{align}
    \frac{(1 - \alpha)^2}{1 - \alpha^2 \gamma} < \frac{\eta(1 - \alpha)}{\eta - \alpha \gamma}.
\end{align}
Since $\alpha^2\gamma<1$ and $\eta>\alpha\gamma$, after some algebra, this reduces to
\begin{align}
    \gamma (\alpha - 1) < \eta (1 - \alpha \gamma).
\end{align}
Because $\alpha \leq 1$, the left-hand side is non-positive, and since $\alpha < \gamma^{-1}$, the right-hand side is strictly positive. Thus this inequality always holds, and the bias is decreasing.

In the case of variance, for the variance to be decreasing, we must have
\begin{align}
    \frac{\alpha^2 \gamma}{1 - \alpha^2 \gamma} < \frac{\alpha \gamma}{\eta - \alpha \gamma}.
\end{align}
Again since $\alpha^2\gamma<1$ and $\eta>\alpha\gamma$, this reduces to
\begin{align}
    \alpha \eta < 1.
\end{align}
So, unless both $\alpha = 1$ and $\eta = 1$, in which case every member of the ensemble is the ordinary least squares predictor fit using the entire training data, the variance is decreasing. 

\subsection{Dropout and Ridge Regression}
\label{sec:dropout}

There is an interesting connection between the ordinary least squares ensemble with $\eta=1$ and the popular \emph{dropout} technique \citep{Srivastava2014dropout} used in deep (neural) network training, which consists of randomly masking the features at each iteration of (stochastic) gradient descent. To draw this connection, define
\begin{align}
    \ell_i(\bbetaprime{}) = \norm[2]{\mX \mS_i \mS_i^\transp \bbetaprime - \vy}^2. 
\end{align}
Then our ensemble member parameter estimates are minimizers of this loss function.
\begin{align}
    \bbetahat{}^{(i)} = \argmin_{\bbetaprime{}} \ell_i(\bbetaprime{}) \text{ s.t. } 
    \bbeta_{S_i^c}' &= {\bm 0}.
\end{align}
For each $i$, the $i$-th member of the ensemble is able to solve its subproblem independently of the other members. As a result, we can consider the ensemble to be a model with $\sum_{i=1}^k |S_i|$ parameters that are eventually averaged to reduce them down to $p$ parameters. If we were to instead constrain ourselves so that we were allowed to use \emph{only} $p$ parameters, such that we could not optimize each member of the ensemble independently, we might try to optimize them jointly by minimizing the average loss. That is,
\begin{align}
    \bbetahat{} = \argmin_{\bbetaprime{}} \frac{1}{k} \sum_{i=1}^k \ell_i(\bbetaprime{}).
\end{align}
If we go a step further and let $k \to \infty$ and optimize this loss using stochastic gradient descent where at each iteration we use the gradient of an individual $\ell_i$ selected at random, then our ensemble becomes equivalent to the predictor learned using dropout. It is well-known that dropout with linear regression has a very strong connection to ridge regression \citep{Srivastava2014dropout}; specifically, we find that
\begin{align}
    \label{eq:dropout}
    \bbetahat{} = \frac{1}{\alpha} \inv{\mX^\transp \mX + \frac{1 - \alpha}{\alpha} \mathrm{diag}(\mX^\transp \mX)} \mX^\transp \vy.
\end{align}
In the case of $\bSigma = \mI_p$, $n^{-1} \mathrm{diag}(\mX^\transp \mX)$ will converge to $\mI_p$ as $n, p \to \infty$, in which case dropout and ridge regression are equivalent up to a rescaling. We discuss the case where $\bSigma \neq \mI_p$ in Section~\ref{sec:non-identity}.

\subsection{Scaled Ensembles}
\label{sec:scaled-ensemble}

\begin{figure}[t]
    \centering
    \includegraphics{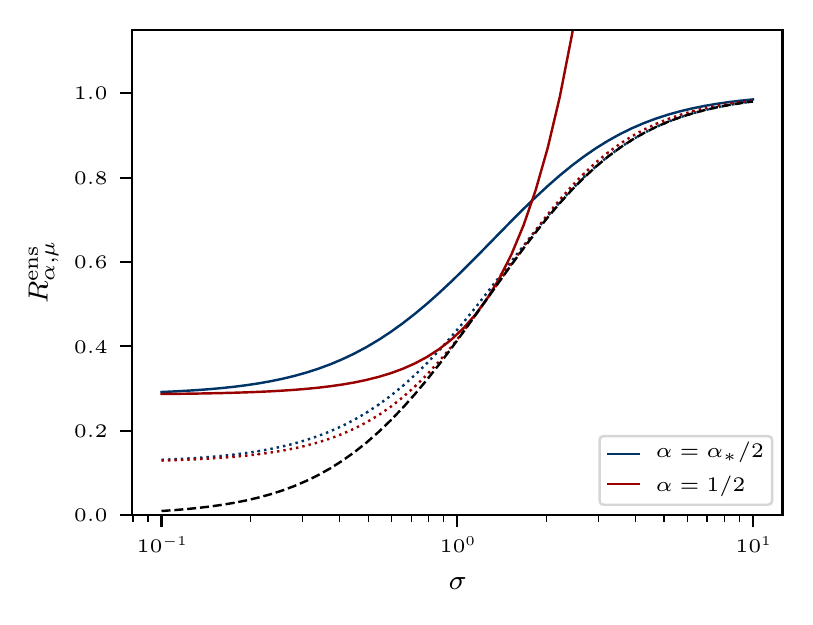}
    \caption{$\mu$-scaled large-ensemble risk (theoretical, $\gamma=0.5$) when using $\mu = 1$ (solid) and $\mu=\mu_*$ (dotted). For both the setting where we use fewer features than optimal with $\alpha=\alpha_*/2$ (blue) and the fixed $\alpha = 1/2$ setting (red), we see significantly improved risk by scaling.}
    \label{fig:tuned-mu}
\end{figure}

Our ensemble combines the individual predictors by simple averaging. However, in light of the fact that dropout is only equivalent to ridge regression up to a rescaling of the output, it is worth considering the effect of using an equally-weighted linear combination but using different weights from $1/k$ in constructing the ensemble predictor. 
That is, 
we consider the risk of the
\emph{$\mu$-scaled} predictor $\bbetahat{}^\ens_{\mu} = (\mu/k) \sum_{i=1}^k \bbetahat{}^{(i)}$. 
 A simple calculation, proved in the Appendix, shows that under the assumptions of Theorem~\ref{thm:ensemble-risk} the large-ensemble risk of the $\mu$-scaled predictor is given by 
\begin{align}
    \label{eq:mu-scaled-risk}
    R_{\alpha, \mu}^\ens{} =  \mu^2R_{\alpha}^\ens{} + (1-\mu)^2 + 2\mu(1-\mu)(1-\alpha). 
\end{align}
Hence, it is possible to minimize the risk of
$\bbetahat{}^\ens_{\mu}$ over
the choice of parameter $\mu$. This results in 
\begin{align}
\mu_* &= \frac{\alpha}{R_{\alpha}^\ens{} + 2\alpha - 1}
\end{align}
as the optimal choice for $\mu$ and 
\begin{align}
    R_{\alpha, \mu_*}^\ens{} &= 1 - \frac{\alpha^2}{2\alpha-1 + R_{\alpha}^\ens{}}
\end{align}
as the achieved risk for the optimally-scaled ensemble. Note that 
as a result of Corollary \ref{corollary:one-minus-alpha}, $R_{\alpha_*}^\ens{} = 1-\alpha_*$. Therefore, for
ensembles with optimally-tuned 
$\alpha = \alpha_*$ we have $\mu_* = 1$, and any scaling in constructing the ensemble predictor will not further improve 
the achieved risk. However, it is easy to see that when 
$\alpha > \alpha_*$ (the ensemble members select \emph{more} features than is optimal), $\mu_* < 1$, and the risk is improved
by adding extra shrinkage to the ensemble predictor. Similarly,
if $\alpha < \alpha_*$, (the ensemble members select \emph{less} features than is optimal), $\mu_* > 1$, and the risk is improved
by inflating the ensemble predictor. We illustrate the improvement in risk to be had in Figure~\ref{fig:tuned-mu}, where we plot the risk with ($\mu=\mu_*$) and without ($\mu=1$) optimal scaling for two choices of $\alpha$---one where we always select half as many features as optimal ($\alpha = \alpha_*/2)$, and one where we always use half of the available features $(\alpha = 1/2)$.

\section{FUTURE DIRECTIONS}

\subsection{Non-Identity Covariance}
\label{sec:non-identity}

Of course, it is important to understand the behavior of the ordinary least squares ensemble in the case where $\bSigma \neq \mI_p$ when considering applications of the method to real data. As discussed in Section~\ref{sec:ensemble-risk}, provided $\bSigma$ is invertible, ridge regression remains the optimal linear predictor, and whether the ensemble (or extensions thereto) still achieves the optimal risk in this setting remains an open question. 

By inspection of the closed-form solution of dropout in \eqref{eq:dropout}, we see that it is no longer equivalent (as $n, p \to \infty$) to ridge regression in this setting and is therefore no longer optimal. We believe that this is likely the case for the ensemble as well. However, if we extend the coin-flipping strategy for feature subset selection to one where we have a collection of coin with probabilities $\balpha \in [0, 1]^p$, one for each feature, we can extend the result in \eqref{eq:dropout} to obtain the closed-form dropout solution
\begin{align}
    \label{eq:generalized-dropout}
    \bbetahat{} = \mA^{-1} \inv{\mX^\transp \mX + \left(\mI_p - \mA\right) \mA^{-1} \mathrm{diag}(\mX^\transp \mX)} \mX^\transp \vy,
\end{align}
where $\mA = \mathrm{diag}(\balpha)$. We prove this result in the Appendix. Thus, if $\balpha$ is chosen such that 
\begin{align}
    \frac{1 - \alpha_j}{\alpha_j} = \frac{\lambda}{n [\bSigma]_{jj}},
\end{align}
then the corrected dropout estimator
\begin{align}
    \widetilde{\bbeta} = \mA \bbetahat{}
\end{align}
is equivalent to ridge regression with parameter $\lambda$ as $n, p \to \infty$. This leads us to believe that the optimal ensemble in the $\bSigma \neq \mI_p$ setting should also use non-uniform feature sampling, and extending our analysis to this case is an interesting area for future work.

\subsection{Beyond Ordinary Least Squares: Ensembles of Interpolators}
\label{sec:fat-submatrices}

Throughout this work we have assumed that the members of the ensemble solve their subproblems using ordinary least squares, which yields the unique solution that minimizes the squared error given $|T_i|$ observations of $|S_i|$ variables, and this uniqueness requires that $|T_i|$ be no less than $|S_i|$. In the case where $|T_i| < |S_i|$, there are infinitely many solutions that minimize the squared error. However, we could in this case opt to \emph{regularize} the solution to solve this problem. While analysis of the effect of regularizing the solution of the subproblems in the ensemble is beyond the scope of this work, we comment briefly on what would happen if we were to simply use the same solution presented in \eqref{eq:pseudoinverse-solution}---i.e., use the pseudoinverse solution, which has the smallest $\ell^2$ norm of all solutions to the least squares problem. In this case, when $\eta = 1$, the learned predictor would be an \emph{interpolator} \citep{belkin2018interpolate, hastie2019surprises} of the training data, and such methods have recently become increasingly of interest given the ability of deep (neural) network methods to have extremely good test performance while having (nearly) zero training error \citep{Zhang2017understanding, Belkin15849}.

Specifically, it becomes immediately clear that in this setting, the effect of the choice of $\eta$ does not vanish as $k \to \infty$. Lemma~\ref{lemma:variance} can easily be extended to this setting, since the roles of $S_i$ and $T_i$ in \eqref{eq:variance-ij} can simply be reversed,
and as $n, p \to \infty$, we obtain
\begin{align}
    \label{eq:variance-finite-alt}
    &\E_{\mX, \setS, \setT} \left[ \variance_{ij}(\bbetahat{}^\ens) \right]
    \xrightarrow{a.s.}
    \begin{cases}
    \frac{\sigma^2 \eta^2}{\gamma - \eta^2} & \text{if } i \neq j, \\
    \frac{\sigma^2 \eta}{\alpha \gamma - \eta} & \text{if } i = j.
    \end{cases}
\end{align}
Thus, the variance component of the large-ensemble risk in this setting is equal to $\sigma^2 \eta^2 / (\gamma - \eta^2)$ and does not depend upon $\alpha$. In future work, we plan to extend our analysis for the bias component of the large-ensemble risk to this setting, and we expect that in this case the bias will depend on \emph{both} $\alpha$ and $\eta$.

\subsection{Optimal Ensemble Mixing}

In the ordinary least squares ensemble, we have used equal weighting when taking the average of our predictors. Instead, we could extend the idea presented in Section~\ref{sec:scaled-ensemble} to consider unequal weighting parameterized by $\bmu \in \reals^k$, giving us the ensemble parameter estimate $\bbetahat{}_\bmu^\ens = \sum_{i=1}^k \mu_i \bbetahat{}^{(i)}$. While equal weighting gives us optimal risk in the setting where $\bbeta \sim \mathcal{N}({\bm 0}, p^{-1} \mI_p)$, where ridge regression is optimal, under other distributional assumptions on $\bbeta$, such as sparsity, where ridge regression is not optimal, unequal weighting has the potential to yield better ensembles.

Using the sparsity example, consider $\bbeta$ such that $\norm[0]{\bbeta} = s \ll p$, and suppose that for some $i$, $S_i = S_\bbeta$, where $S_\bbeta = \set{j : \beta_j \neq 0}$. For simplicity, assume that $\eta = 1$, so that $T_i = [n]$ for all $i$. In this case, any predictor that uses the remaining $p-s$ features injects noise into its predictions, so the best predictor uses only the $s$ features in $S_\bbeta$. Under the i.i.d.\ Gaussian noise assumption, the predictor with lowest risk is in fact
\begin{align}
    \bbetahat &= \argmin_{\bbetaprime{} : \bbeta_{S_\bbeta^c}' = {\bm 0}}
    \norm[2]{\vy - \mX \bbetaprime{}} = \bbetahat{}^{(i)},
\end{align}
where $i$ is such that $S_i = S_\bbeta$. Thus an optimal weighting $\bmu$ is given by
\begin{align}
    \mu_i = \begin{cases}
    \frac{1}{C} & \text{if } S_i = S_\bbeta, \\
    0 & \text{otherwise},
    \end{cases}
\end{align}
where $C = |\set{i : S_i = S_\bbeta}|$. This optimal weighting is decidedly non-uniform, and this raises the question of what schemes could be employed, either adaptively or non-adaptively, to minimize risk, and how they would fit into this analysis framework. 

\subsubsection*{Acknowledgements}
We would like to thank Ryan Tibshirani for helpful discussions and the anonymous reviewers for their helpful feedback.
This work was supported by 
NSF grants CCF-1911094, IIS-1838177, and IIS-1730574; 
ONR grants N00014-18-12571 and N00014-17-1-2551;
AFOSR grant FA9550-18-1-0478; 
DARPA grant G001534-7500; and a 
Vannevar Bush Faculty Fellowship, ONR grant N00014-18-1-2047.

\balance

\clearpage
\onecolumn
\appendix
\section{USEFUL LEMMAS}

The following two lemmas will be useful in deriving the bias and variance terms of the ensemble risk. Their proofs can be found in Section~\ref{sec:lemma-proofs}.

\begin{lemma}
    \label{lemma:subselect-smaller}
    Let $S \subseteq [p]$ be a subset with corresponding selection matrix $\mS$, and let $\mS^c$ be the selection matrix corresponding to $S^c$. Then for a random matrix $\mX \in \reals^{n \times p}$ with rows independently drawn from $\mathcal{N}(\bm{0}, \mI_p)$ such that $n > |S|$, and for any random function $f : \reals^{n \times |S|} \to \reals^{n \times |S|}$ that $f(\mX \mS)$ and $\mX \mS^c$ are independent,
    \begin{align}
        \E_{\mX \mS^c} \left[
        \mS^\transp \mX^\dagger
        \right]
        = \left(
        \mX \mS
        \right)^\dagger
    \end{align}
    and
    \begin{align}
        \E_{\mX \mS^c} \left[
        {\mS^c}^\transp \mX^\transp f(\mX \mS)
        \mS^\transp \mX^\dagger
        \right]
        = \bf{0}.
    \end{align}
\end{lemma}

\begin{lemma}
    \label{lemma:subselect-larger}
    Let $T_1, T_2 \subseteq [n]$ be independent random subsets with corresponding selection matrices $\mT_1, \mT_2$ such that $
    \E \left[ \mT_j \mT_j^\transp \right] = \frac{|T_j|}{n} \mI_n
    $. Then for random matrices $\mX \in \reals^{n \times p_X}, \mY \in \reals^{n \times p_Y}$ independent of $T_1$ and $T_2$ with independent and identically distributed rows such that $\mX^\transp \mT_j \mT_j^\transp \mX$ and $\mY^\transp \mT_j \mT_j^\transp \mY$ are invertible, and for any matrix $\mA \in \reals^{p_X \times p_Y}$,
    \begin{align}
        \E_{T_1, T_2} \left[
        \left( \mT_1^\transp \mX \right)^\dagger \mT_1^\transp 
        \left( \left( \mT_2^\transp \mX \right)^\dagger \mT_2^\transp \right)^\transp
        \right]
        = 
        \left(\mX^\transp \mX \right)^\dagger
    \end{align}
    and
    \begin{align}
        \E_{T_1, T_2} \left[
        \left( \left( \mT_1^\transp \mX \right)^\dagger \mT_1^\transp \right)^\transp
        \mA
        \left( \mT_2^\transp \mY \right)^\dagger
        \mT_2^\transp 
        \right]
        = 
        \left(\mX^\dagger \right)^\transp
        \mA
        \mY^\dagger .
    \end{align}
\end{lemma}

\section{PROOF OF LEMMA~\ref{lemma:bias} (BIAS)}

To compute the bias, we need to evaluate terms of the form
\begin{align}
    \E_{\mX, \setS, \setT} \left\langle
    \bbeta \bbeta^\transp, 
    \left(
    \mI_p - \mS_i \left(\mT_i^\transp \mX \mS_i\right)^\dagger \mT_i^\transp \mX 
    \right)^\transp
    \left(
    \mI_p - \mS_j \left(\mT_j^\transp \mX \mS_j\right)^\dagger \mT_j^\transp \mX 
    \right)
    \right\rangle.
\end{align}
First, we note that since $\mS_i \mS_i^\transp + \mS_i^c {\mS_i^c}^\transp = \mI_p$,
\begin{align}
    \mI_p - \mS_i \left(\mT_i^\transp \mX \mS_i\right)^\dagger \mT_i^\transp \mX
    &= \mI_p - \mS_i \left(\mT_i^\transp \mX \mS_i\right)^\dagger \mT_i^\transp \mX \left( \mS_i \mS_i^\transp + \mS_i^c {\mS_i^c}^\transp \right) \\
    &= \mI_p - \mS_i \mS_i^\transp -  \mS_i \left(\mT_i^\transp \mX \mS_i\right)^\dagger \mT_i^\transp \mX \mS_i^c {\mS_i^c}^\transp \\
    &= \left(\mI_p -  \mS_i \left(\mT_i^\transp \mX \mS_i\right)^\dagger \mT_i^\transp \mX
    \right)
    \mS_i^c {\mS_i^c}^\transp.
\end{align}
So, we can equivalently evaluate
\begin{align}
    \E_{\mX, \setS, \setT} \left\langle
    \bbeta\bbeta^\transp,
    \mS_i^c {\mS_i^c}^\transp
    \left[
    \mI_p - 
    \mX^\transp \mT_i 
    \left(\mS_i^\transp \mX^\transp \mT_i \right)^\dagger
    \mS_i^\transp
    \right]
    \left[
    \mI_p - 
    \mS_j \left(\mT_j^\transp \mX \mS_j \right)^\dagger 
    \mT_j^\transp \mX
    \right]
    \mS_j^c {\mS_j^c}^\transp 
    \right\rangle.
\end{align}

It suffices to evaluate the expectation of the second argument of the inner product:
\begin{align}
    \E_{\mX, \setS, \setT} \bigg[
    &\mS_i^c {\mS_i^c}^\transp
    \left[
    \mI_p - 
    \mX^\transp \mT_i 
    \left(\mS_i^\transp \mX^\transp \mT_i \right)^\dagger
    \mS_i^\transp
    \right]
    \left[
    \mI_p - 
    \mS_j \left(\mT_j^\transp \mX \mS_j \right)^\dagger 
    \mT_j^\transp \mX
    \right]
    \mS_j^c {\mS_j^c}^\transp 
    \bigg] \nonumber \\
    ={} &\E_{\mX, \setS, \setT} \bigg[
    \mS_i^c {\mS_i^c}^\transp
    \mX^\transp \mT_i 
    \left(\mS_i^\transp \mX^\transp \mT_i \right)^\dagger
    \mS_i^\transp
    \mS_j \left(\mT_j^\transp \mX \mS_j \right)^\dagger 
    \mT_j^\transp \mX
    \mS_i^\transp
    \mS_j^c {\mS_j^c}^\transp \nonumber \\
    &- \mS_i^c {\mS_i^c}^\transp
    \mX^\transp \mT_i 
    \left(\mS_i^\transp \mX^\transp \mT_i \right)^\dagger
    \mS_i^\transp - \mS_j \left(\mT_j^\transp \mX \mS_j \right)^\dagger 
    \mT_j^\transp \mX
    \mS_j^c {\mS_j^c}^\transp 
    + \mS_i^c {\mS_i^c}^\transp \mS_j^c {\mS_j^c}^\transp \bigg].
\end{align}
The second and third terms are zero in expectation. To see this for the second term, observe that ${\mS_i^c}^\transp \mX^\transp$ and $\mS_i^\transp \mX^\transp$ are independent and each zero-mean. An analogous argument applies to the third term. The fourth term is equal to
\begin{align}
\frac{|S_i^c \cap S_j^c|}{p} \mI_p.
\label{eq:bias:fourth-term}
\end{align}

We now consider the case where $i \neq j$. To evaluate the first term, we first apply Lemma~\ref{lemma:subselect-larger}. This simplifies the expression to
\begin{align}
    \E_{\mX, \setS} \left[
    \mS_i^c {\mS_i^c}^\transp
    \mX^\transp
    \left(\mS_i^\transp \mX^\transp \right)^\dagger
    \mS_i^\transp
    \mS_j \left( \mX \mS_j \right)^\dagger 
    \mX
    \mS_i^\transp
    \mS_j^c {\mS_j^c}^\transp
    \right].
    \label{eq:bias:first-term}
\end{align}
Now let $\mS_{i \cap j}$, $\mS_{i \setminus j}$, $\mS_{j \setminus i}$, and $\mS_{i \cup j}^c$ denote the selection matrices corresponding to the sets $S_i \cap S_j$, $S_i \setminus S_j$, $S_j \setminus S_i$, and $S_i^c \cap S_j^c$, respectively. Without loss of generality, consider when $\mS_i^c = \begin{bmatrix}\mS_{j \setminus i} \; \mS_{i \cup j}^c \end{bmatrix}$ and $\mS_j^c = \begin{bmatrix}\mS_{i \setminus j} \; \mS_{i \cup j}^c \end{bmatrix}$. Then the matrix inside this expectation is of the form
\begin{align}
    \mS_i^c \begin{bmatrix}
    \mA & \mB \\
    \mC & \mD
    \end{bmatrix} {\mS_j^c}^\transp,
\end{align}
where
\begin{align}
    \mA &= \mS_{j \setminus i}^\transp
    \mX^\transp
    \left(\mS_i^\transp \mX^\transp \right)^\dagger
    \mS_i^\transp
    \mS_j \left( \mX \mS_j \right)^\dagger 
    \mX
    \mS_{i \setminus j} \\
    \mB &= \mS_{j \setminus i}^\transp
    \mX^\transp
    \left(\mS_i^\transp \mX^\transp \right)^\dagger
    \mS_i^\transp
    \mS_j \left( \mX \mS_j \right)^\dagger 
    \mX
    \mS_{i \cup j}^c \\
    \mC &= {\mS_{i \cup j}^c}^\transp
    \mX^\transp
    \left(\mS_i^\transp \mX^\transp \right)^\dagger
    \mS_i^\transp
    \mS_j \left( \mX \mS_j \right)^\dagger 
    \mX
    \mS_{i \setminus j} \\
    \mD &= {\mS_{i \cup j}^c}^\transp
    \mX^\transp
    \left(\mS_i^\transp \mX^\transp \right)^\dagger
    \mS_i^\transp
    \mS_j \left( \mX \mS_j \right)^\dagger 
    \mX
    \mS_{i \cup j}^c.
\end{align}
In the case of $\mB$ and $\mC$, because $\mX \mS_{i \cup j}^c$ is independent of the remainder of the factors, $\E_\mX \left[ \mB \right]$ and $\E_\mX \left[ \mC \right]$ are equal to $\bm{0}$. By applying the second claim of Lemma~\ref{lemma:subselect-smaller}, we observe that $\E_\mX \left[ \mA \right]$ is also equal to $\bm{0}$. This leaves
\begin{align}
    \E_\mX \left[ \mD \right] 
    &= \E_\mX \left[
    {\mS_{i \cup j}^c}^\transp
    \mX^\transp
    \E_{\mX \mS_{j \setminus i}} \left[
    \left(\mS_i^\transp \mX^\transp \right)^\dagger
    \mS_i^\transp
    \mS_{i \cap j}
    \right]
    \E_{\mX \mS_{j \setminus i}} \left[ 
    \mS_{i \cap j}^\transp
    \mS_j \left( \mX \mS_j \right)^\dagger 
    \right]
    \mX
    \mS_{i \cup j}^c
    \right] \\
    &= \E_\mX \left[
    {\mS_{i \cup j}^c}^\transp
    \mX^\transp
    \left( \mX \mS_{i \cap j} \mS_{i \cap j}^\transp \mX^\transp  \right)^\dagger 
    \mX
    \mS_{i \cup j}^c
    \right].
\end{align}
We can evaluate the expectation of the pseudoinverse on its own since $\mX \mS_{i \cap j}$ and $\mX \mS_{i \cup j}^c$ are independent. This matrix has a generalized inverse Wishart distribution with scale matrix $\mI_n$ and $|S_i \cap S_j|$ degrees of freedom, which yields
\begin{align}
    \E_\mX \left[
    \left( \mX \mS_{i \cap j} \mS_{i \cap j}^\transp \mX^\transp  \right)^\dagger
    \right]
    &= \frac{|S_i \cap S_j|}{n(n - |S_i \cap S_j| - 1)} \mI_n.
\end{align}
This leaves
\begin{align}
    \E_\mX \left[
    {\mS_{i \cup j}^c}^\transp
    \mX^\transp
    \left( \frac{|S_i \cap S_j|}{n(n - |S_i \cap S_j| - 1)} \mI_n \right)
    \mX
    \mS_{i \cup j}^c
    \right]
    &= \frac{|S_i \cap S_j|}{n - |S_i \cap S_j| - 1} \mI_{|S_i^c \cap S_j^c|}.
\end{align}
Then the expectation in \eqref{eq:bias:first-term} becomes
\begin{align}
    \E_\setS \left[ \frac{|S_i \cap S_j|}{(n - |S_i \cap S_j| - 1)}
    \mS_i^c \begin{bmatrix}
    \bm{0} & \bm{0} \\
    \bm{0} & \mI_{|S_i^c \cap S_j^c|}
    \end{bmatrix} {\mS_j^c}^\transp
    \right]
    &= \frac{|S_i \cap S_j||S_i^c \cap S_j^c|}{p(n - |S_i \cap S_j| - 1)} \mI_p,
\end{align}
and combing with \eqref{eq:bias:fourth-term}, we have that the bias is equal to
\begin{align}
    \frac{|S_i^c \cap S_j^c|}{p}
    \left(1 + \frac{|S_i \cap S_j|}{n - |S_i \cap S_j| - 1} 
    \right)
    \norm[2]{\bbeta}^2.
\end{align}

When $i = j$, by a similar argument, without the need to apply Lemma~\ref{lemma:subselect-larger}, it follows that the bias is equal to
\begin{align}
    \frac{|S_i^c|}{p}
    \left(1 + \frac{|S_i|}{|T_i| - |S_i| - 1} 
    \right)
    \norm[2]{\bbeta}^2.
\end{align}

\section{PROOF OF LEMMA~\ref{lemma:variance} (VARIANCE)}
To compute the variance, we need to evaluate the terms of the form
\begin{align}
    \E_{\mX, \setT} \left\langle
    \mS_i \left(
    \mT_i^\transp \mX \mS_i
    \right)^\dagger
    \mT_i^\transp,
    \mS_j \left(
    \mT_j^\transp \mX \mS_j
    \right)^\dagger
    \mT_j^\transp
    \right\rangle.
\end{align}
Let $\mS$ be the selection matrix corresponding to the set $S_i \cap S_j$. Then
\begin{align}
    &\E \left\langle
    \mS_i \left(
    \mT_i^\transp \mX \mS_i
    \right)^\dagger
    \mT_i^\transp,
    \mS_j \left(
    \mT_j^\transp \mX \mS_j
    \right)^\dagger
    \mT_j^\transp
    \right\rangle \nonumber \\
    &= \E \left\langle
    \mS^\transp \mS_i \left(
    \mT_i^\transp \mX \mS_i
    \right)^\dagger
    \mT_i^\transp,
    \mS^\transp \mS_j \left(
    \mT_j^\transp \mX \mS_j
    \right)^\dagger
    \mT_j^\transp
    \right\rangle \\
    &= \E \left\langle
    \E_{\mX^{S_i \setminus S_j}} \left[
    \mS^\transp \mS_i \left(
    \mT_i^\transp \mX \mS_i
    \right)^\dagger
    \right]
    \mT_i^\transp,
    \E_{\mX^{S_j \setminus S_i}} \left[
    \mS^\transp \mS_j \left(
    \mT_j^\transp \mX \mS_j
    \right)^\dagger
    \right]
    \mT_j^\transp
    \right\rangle \\
    \label{eq:variance-proof:apply-lemma:subselect-smaller}
    &= \E \left\langle
    \left(
    \mT_i^\transp \mX \mS
    \right)^\dagger
    \mT_i^\transp,
    \left(
    \mT_j^\transp \mX \mS
    \right)^\dagger
    \mT_j^\transp
    \right\rangle.
\end{align}
The equality~\eqref{eq:variance-proof:apply-lemma:subselect-smaller} is the result of two applications of Lemma~\ref{lemma:subselect-smaller}. 

In the case that $i \neq j$, an application of Lemma~\ref{lemma:subselect-larger} simplifies the above to
\begin{align}
    \tr \left( \E_\mX \left[
    \inv{\mS^\transp \mX^\transp \mX \mS}
    \right] \right)
    = \frac{|S_i \cap S_j|}{n - |S_i \cap S_j| - 1}.
\end{align}
The equality comes from $\inv{\mS^\transp \mX^\transp \mX \mS}$ having an inverse Wishart distribution with scale matrix $\mI_{|S_i \cap S_j|}$ and $n$ degrees of freedom. 

When $i = j$, we obtain a similar result directly without needing Lemma~\ref{lemma:subselect-larger}. The above simplifies to
\begin{align}
    \tr \left( \E_\mX \left[
    \inv{\mS_i^\transp \mX^\transp \mT_i \mT_i^\transp \mX \mS_i}
    \right] \right)
    = \frac{|S_i|}{|T_i| - |S_i| - 1}.
\end{align}

\section{PROOF OF THEOREM~\ref{thm:like-ridge}}

We first introduce the result due to \citet{dobriban2018}. We note again, as we noted in the main text, that in the setting of $\bSigma = \mI_p$, where the optimal ridge regression risk is equal to the estimation error of the minimum mean squared error (MMSE) estimator, results on the value of this quantity predate the result of \citet{dobriban2018}. We refer the reader, for example, to the wireless communication literature \citep[see, e.g.,][]{tulino2004random}. However, \citet{dobriban2018} have developed the first results on ridge regression risk for general $\bSigma$, and their clean theorem statement is simple and straightforward to use, even in the $\bSigma = \mI_p$ case.
\begin{proposition}[{from \citealp[Theorem~2.1]{dobriban2018}}]
Assume that $\Sigma = \mI_p$ and $\bbeta \sim \mathcal{N}({\bm 0}, p^{-1} \mI_p)$. Then in the limit as $n, p \to \infty$ with $p/n \to \gamma$, we have almost surely that
\begin{align}
    \inf_\lambda R(\bbetahat{}_\lambda^\ridge)
    = \frac{1}{2} \left( 
    \frac{\gamma - 1}{\gamma} - \sigma^2
    + \sqrt{
    \left(\sigma^2 - \frac{\gamma - 1}{\gamma} \right)^2 + 4 \sigma^2
    }
    \right).
\end{align}
\end{proposition}
We note that this expression is equal to $\sigma^2 (R^*(1/\sigma^2, \gamma) - 1)$ in the notation of \citet{dobriban2018}, where this transformation is necessary because we assume $\norm[2]{\bbeta} = 1$ rather than $\sigma = 1$ and because we evaluate the noise-free risk.

The minimizer of the large ensemble risk should satisfy the first-order optimality condition, so we begin by taking its derivative.
\begin{align}
    \frac{dR_\alpha^\ens}{d\alpha} 
    &= \frac{
    (-2(1-\alpha) + 2\sigma^2 \alpha \gamma) (1 - \alpha^2 \gamma)
    - ((1 - \alpha)^2 + \sigma^2 \alpha^2 \gamma) (-2\alpha \gamma)
    }{
    (1 - \alpha^2 \gamma)^2
    } \\
    \label{eq:risk-derivative}
    &= \frac{
    -\alpha^2 \gamma + (\gamma (\sigma^2 + 1) + 1) - 1
    }{
    (1 - \alpha^2 \gamma)^2
    }.
\end{align}
Thus the minimizer $\alpha_*$ should satisfy
\begin{align}
    \label{eq:alpha-quadratic}
    \alpha_*^2 \gamma - \alpha_* (\gamma (\sigma^2 + 1) + 1) + 1 = 0.
\end{align}
From here, it is simply a matter of cumbersome algebra to show that the choice
\begin{align}
    \label{eq:smaller-alpha}
    \alpha_* = \frac{
        \gamma (\sigma^2 + 1) + 1 
        - \sqrt{(\gamma (\sigma^2 + 1) + 1)^2 - 4\gamma}
    }{2\gamma}
\end{align}
is the valid root of this quadratic expression and is such that $R_{\alpha_*}^\ens = \inf_\lambda R(\bbetahat{}_\lambda^\ridge)$. We here show a slightly more interesting approach, leading to Corollary~\ref{corollary:one-minus-alpha}. First, we start from \eqref{eq:alpha-quadratic} and add a root of $\alpha_*=0$, and then we proceed to manipulate the resulting equation.
\begin{align}
    \alpha_*(\alpha_*^2 \gamma - \alpha_* (\gamma (\sigma^2 + 1) + 1) + 1) &= 0 \\
    \alpha_* - \alpha_*^2 (\gamma (\sigma^2 + 1) + 1) &= - \alpha_*^3 \gamma \\
    2\alpha_* - \alpha_*^2 (\gamma (\sigma^2 + 1) + 1) &= \alpha_* (1 - \alpha_*^2 \gamma) \\
    \frac{2\alpha_* - \alpha_*^2 (\gamma (\sigma^2 + 1) + 1)}{1 - \alpha_*^2 \gamma} &= \alpha_*.
\end{align}
Continuing from this last equation,
\begin{align}
    \alpha_* 
    &= \frac{2\alpha_* - \alpha_*^2 (\gamma (\sigma^2 + 1) + 1)}{1 - \alpha_*^2 \gamma} \\
    &= \frac{2\alpha_* - \sigma^2 \alpha_*^2 \gamma - \alpha^2 \gamma - \alpha^2 + 1 - 1}{1 - \alpha_*^2 \gamma} \\
    &= \frac{1 - \alpha^2 \gamma - (1 - 2\alpha_* + \alpha_*^2) - \sigma^2 \alpha_*^2 \gamma}{1 - \alpha_*^2 \gamma} \\
    &= 1 - \frac{(1 - \alpha_*)^2 + \sigma^2 \alpha_*^2 \gamma}{1 - \alpha_*^2 \gamma} \\
    &= 1 - R_{\alpha_*}^\ens.
\end{align}
Thus, if $\alpha_*$ is a root of \eqref{eq:alpha-quadratic} or $\alpha_* = 0$, then $R_{\alpha_*}^\ens = 1 - \alpha_*$. We proceed by checking the larger root of \eqref{eq:alpha-quadratic}, but before doing so, we derive the following equality:
\begin{align}
    (\gamma(\sigma^2 + 1) + 1)^2 - 4 \gamma 
    &= (\gamma(\sigma^2 + 1) + 1)^2 - (4 \gamma^2 (\sigma^2 + 1) + 4 \gamma) + 4 \gamma^2 + 4 \sigma^2 \gamma^2 \\
    &= (\gamma(\sigma^2 + 1) + 1 - 2\gamma)^2 + 4 \sigma^2 \gamma^2 \\
    &= (\gamma(\sigma^2 - 1) + 1)^2 + 4\sigma^2 \gamma^2.
\end{align}
Now, we observe for the larger root (which we denote as $\alpha_*'$) that
\begin{align}
    \alpha_*' &= \frac{
        \gamma (\sigma^2 + 1) + 1 
        + \sqrt{(\gamma(\sigma^2 - 1) + 1)^2 + 4\sigma^2 \gamma^2}
    }{2\gamma} \\
    &\geq \frac{1}{2} \left( 
    \sigma^2 + 1 + \frac{1}{\gamma} + \left|\sigma^2 - 1 + \frac{1}{\gamma} \right|
    \right) \\
    &= \begin{cases}
    \sigma^2 + \frac{1}{\gamma} & \text{if } \frac{1}{\gamma} > 1 - \sigma^2 \\
    1 & \text{if } \frac{1}{\gamma} \leq 1 - \sigma^2.
    \end{cases}
\end{align}
Thus the only case where $\alpha_*'$ is a valid hyperparameter choice (that is, $\alpha_* \leq \min \set{1, \gamma^{-1}}$) is when $\sigma^2 = 0$ and $\gamma = 1$, in which case $\alpha_* = 1$ is a double root of \eqref{eq:alpha-quadratic}. So it suffices to evaluate the smaller root even in that case. Now that we know that $\alpha_*'$ is not conatined in $[0, \min \set{1, \gamma^{-1}}]$ (except in the aforementioned special case) and that by inspection of $R_\alpha^\ens$ the asymptote at $\alpha = \gamma^{-1/2}$ is not contained in this interval, if we can show that the smaller root (which we denote simply as $\alpha_*$) of \eqref{eq:alpha-quadratic} is contained in this interval, then we know that it is the minimizer of $R_\alpha^\ens$.

For the smaller root, it is clear from \eqref{eq:smaller-alpha} that $\alpha_* \geq 0$. We show by a series of equivalences that $\alpha_* \leq 1/\gamma$:
\begin{align}
    & \alpha_* = \frac{1}{2} \left( \sigma^2 + 1 + \frac{1}{\gamma} - \sqrt{\left( \sigma^2 - 1 + \frac{1}{\gamma} \right)^2 + 4 \sigma^2} \right) \leq \frac{1}{\gamma} \\
    \iff{} &
    \sigma^2 + 1 - \frac{1}{\gamma} \leq \sqrt{\left( \sigma^2 - 1 + \frac{1}{\gamma} \right)^2 + 4 \sigma^2} \\
    \iff{} &
    \left(\sigma^2 + 1 - \frac{1}{\gamma} \right)^2 \leq \left( \sigma^2 - 1 + \frac{1}{\gamma} \right)^2 + 4 \sigma^2 \\
    \iff{} &
    \left(\sigma^2 - 1 + \frac{1}{\gamma} \right)^2 + 4\sigma^2 - 4 \frac{\sigma^2}{\gamma} \leq \left( \sigma^2 - 1 + \frac{1}{\gamma} \right)^2 + 4 \sigma^2 \\
    \iff{} &
    \left(\sigma^2 - 1 + \frac{1}{\gamma} \right)^2 + 4\sigma^2 - 4 \frac{\sigma^2}{\gamma} \leq \left( \sigma^2 - 1 + \frac{1}{\gamma} \right)^2 + 4 \sigma^2 \\
    \iff{} & 0 \leq \frac{\sigma^2}{\gamma}.
\end{align}
The last inequality is always true. Further, we note that every equivalence here still holds under strict inequalities, so for $\sigma > 0$, we have that $\alpha_* < \gamma^{-1}$. By a similar argument, we can show that $\alpha_* \leq 1$ and that $\alpha_* < 1$ if and only if $\sigma > 0$. By the form of the derivative in \eqref{eq:risk-derivative}, we know that $\alpha_*$, as the smaller root, is a local minimum, and therefore it must be the minimum of $R_\alpha^\ens$ on $[0, \min \set{1, \gamma^{-1}}]$. Evaluating the risk at $\alpha_*$, we have
\begin{align}
    R_{\alpha_*}^\ens &= 
    1 - \alpha_* \\
    &= 1 - \frac{1}{2} \left( \sigma^2 + 1 + \frac{1}{\gamma} - \sqrt{\left( \sigma^2 - 1 + \frac{1}{\gamma} \right)^2 + 4 \sigma^2} \right) \\
    &= \frac{1}{2} \left( 1 - \sigma^2 - \frac{1}{\gamma} + \sqrt{\left( \sigma^2 - 1 + \frac{1}{\gamma} \right)^2 + 4 \sigma^2} \right) \\
    &= \frac{1}{2} \left( \frac{\gamma - 1}{\gamma} - \sigma^2 + \sqrt{\left( \sigma^2 - \frac{\gamma - 1}{\gamma} \right)^2 + 4 \sigma^2} \right) \\
    &= \inf_\lambda R(\bbetahat{}_\lambda^\ridge).
\end{align}

\section{PROOFS OF DISCUSSION RESULTS}

\subsection{Proof of Equation~(\ref{eq:mu-scaled-risk}) ($\mu$-scaled Risk)}

Under the assumption that $\bSigma = \mI_p$, the $\mu$-scaled risk is given by
\begin{align}
    R(\mu \bbetahat{}^\ens{}) &= \norm[2]{\bbeta - \mu \bbetahat{}^\ens{}}^2 \\
    &= \norm[2]{(1 - \mu) \bbeta + \mu(\bbeta - \bbetahat{}^\ens{})}^2 \\
    &= (1 - \mu)^2 \norm[2]{\bbeta}^2 
    + 2 (1 - \mu) \mu \left\langle \bbeta, \bbeta - \bbetahat{}^\ens{} \right\rangle
    + \mu^2 \norm[2]{\bbeta - \bbetahat{}^\ens{}}^2
\end{align}
Examining the inner product, we find that
\begin{align}
    \E_{\mX,\vz,\setS,\setT} \left[
    \left\langle \bbeta, \bbeta - \bbetahat{}^\ens{} \right\rangle
    \right]
    &= 
    \left\langle \bbeta,
    \E_{\mX,\vz,\setS,\setT} \left[
    \bbeta - \bbetahat{}^\ens{}
    \right]
    \right\rangle \\
    \label{eq:mu-scaled-risk-expect-z}
    &= 
    \left\langle \bbeta,
    \E_{\mX,\setS,\setT} \left[
    \mI_p - \frac{1}{k} \sum_{i=1}^k \mS_i \left(\mT_i^\transp \mX \mS_i\right)^\dagger \mT_i^\transp \mX
    \right]
    \bbeta
    \right\rangle,
\end{align}
where the equation \eqref{eq:mu-scaled-risk-expect-z} holds because $\E[\vz] = {\bm 0}$. Because the subsamplings are identically distributed, we have
\begin{align}
    \E_{\mX,\setS,\setT} \left[
    \mI_p - \frac{1}{k} \sum_{i=1}^k \mS_i \left(\mT_i^\transp \mX \mS_i\right)^\dagger \mT_i^\transp \mX
    \right]
    &= 
    \mI_p - \E_{\mX,\setS,\setT} \left[
    \mS_i \left(\mT_i^\transp \mX \mS_i\right)^\dagger \mT_i^\transp \mX
    \right] \\
    &=
    \mI_p - \E_{\mX,\setS,\setT} \left[
    \mS_i \left(\mT_i^\transp \mX \mS_i\right)^\dagger \mT_i^\transp 
    \mX \left(\mS_i \mS_i^\transp + \mS_i^c {\mS_i^c}^\transp \right)
    \right] \\
    \label{eq:mu-scaled-risk-expect-x}
    &=
    \mI_p - \E_{\setS} \left[
    \mS_i \left(\mT_i^\transp \mX \mS_i\right)^\dagger \mT_i^\transp 
    \mX \mS_i \mS_i^\transp
    \right] \\
    &=
    \mI_p - \E_{\setS} \left[
    \mS_i \mS_i^\transp
    \right] \\
    &= (1 - \alpha)\mI_p,
\end{align}
where the equation \eqref{eq:mu-scaled-risk-expect-x} holds because $\E[\mX \mS_i^c] = {\bm 0}$. Thus
\begin{align}
    \E_{\mX,\vz,\setS,\setT} [R(\mu \bbetahat{}^\ens{})]
    &= (1 - \mu)^2 \norm[2]{\bbeta}^2 
    + 2 (1 - \mu) \mu 
    \E_{\mX,\vz,\setS,\setT} \left[\left\langle \bbeta, \bbeta - \bbetahat{}^\ens{} \right\rangle \right]
    + \mu^2 \E_{\mX,\vz,\setS,\setT} \left[ \norm[2]{\bbeta - \bbetahat{}^\ens{}}^2 \right] \\
    &= 
    (1 - \mu)^2 + 2 (1 - \mu) \mu (1 - \alpha) + \mu^2 R_\alpha^\ens,
\end{align}
where the last equality holds because $\langle \bbeta, \bbeta \rangle = \norm[2]{\bbeta}^2 = 1$.

\subsection{Proof of Equation~(\ref{eq:generalized-dropout}) (Generalized Dropout)}

For $k \to \infty$, dropout minimizes the expected loss:
\begin{align}
    \E_{S_i} \left[
    \ell_i(\bbetaprime)
    \right]
    &= 
    \E_{S_i} \left[
    \norm[2]{\mX \mS_i \mS_i^\transp \bbetaprime - \vy}^2
    \right].
\end{align}
The expected loss is convex in $\bbetaprime$, so we can find its minimizer by the first order optimality condition:
\begin{align}
    \nabla_{\bbetaprime} \E_{S_i} \left[
    \ell_i(\bbetaprime)
    \right] 
    = 
    \E_{S_i} \left[
    \mS_i \mS_i^\transp \mX^\transp \left( \mX \mS_i \mS_i^\transp \bbetaprime - \vy \right)
    \right]
    = 0
\end{align}
Thus,
\begin{align}
    \bbetahat{} = \inv{ 
    \E_{S_i} \left[
    \mS_i \mS_i^\transp \mX^\transp \mX \mS_i \mS_i^\transp
    \right]
    } 
    \E_{S_i} \left[
    \mS_i \mS_i^\transp \right] \mX^\transp \vy.
\end{align}
Turning first to the inverse, consider that
\begin{align}
    \left[ \E_{S_i} \left[
    \mS_i \mS_i^\transp \mX^\transp \mX \mS_i \mS_i^\transp
    \right] \right]_{j\ell} 
    &= 
    \left[ \mX^\transp \mX \right]_{j \ell} \Pr(j \in S_i, \ell \in S_i),
\end{align}
and that
\begin{align}
    \Pr(j \in S_i, \ell \in S_i)
    &= \begin{cases}
    \alpha_j & \text{if } j = \ell, \\
    \alpha_j \alpha_\ell & \text{otherwise}.
    \end{cases}
\end{align}
This gives us
\begin{align}
    \E_{S_i} \left[
    \mS_i \mS_i^\transp \mX^\transp \mX \mS_i \mS_i^\transp
    \right]
    &= \mA \mX^\transp \mX \mA + \mathrm{diag}(\mX^\transp \mX) (\mA - \mA^2),
\end{align}
where $\mA = \mathrm{diag}(\balpha)$. By a similar and simpler argument,
\begin{align}
    \E_{S_i} \left[
    \mS_i \mS_i^\transp \right] = \mA,
\end{align}
which all together yields
\begin{align}
    \bbetahat{} &= \inv{ 
    \mA \mX^\transp \mX \mA + \mathrm{diag}(\mX^\transp \mX) (\mI_p - \mA) \mA
    } 
    \mA \mX^\transp \vy \\
    &= \mA^{-1} \inv{
    \mX^\transp \mX + \mA^{-1} \mathrm{diag}(\mX^\transp \mX) (\mI_p - \mA)
    } \mX^\transp \vy.
\end{align}

\section{PROOFS OF LEMMAS~\ref{lemma:subselect-smaller} and \ref{lemma:subselect-larger}}
\label{sec:lemma-proofs}

\subsection{Proof of Lemma~\ref{lemma:subselect-smaller}}

Without loss of generality, let $\begin{bmatrix} \mX_1 \; \mX_2 \end{bmatrix} = \mX$, such that $\mX_2 = \mX \mS$. Let $\begin{bmatrix}\mY_1 \\ \mY_2\end{bmatrix} = \mX^\dagger$ be a partitioning of the pseudo-inverse of $\mX$ in the same manner, such that $\mY_2 = \mS^\transp \mX^\dagger = \mS^\transp \inv{\mX^\transp \mX} \mX^\transp$. Then the Gram matrix can be written as
\begin{align}
    \mX^\transp \mX = \begin{bmatrix}
        \mX_1^\transp \mX_1 & \mX_1^\transp \mX_2 \\
        \mX_2^\transp \mX_1 & \mX_2^\transp \mX_2
    \end{bmatrix},
\end{align}
and using block matrix inversion, the inverse admits the form $\inv{\mX^\transp \mX} = \begin{bmatrix} \mA & \mB \\ \mC & \mD \end{bmatrix}$. The relevant quantities are
\begin{align}
    \mC &= - \mD \mX_2^\transp \mX_1 \inv{\mX_1^\transp \mX_1} \\
    \mD &= \inv{\mX_2^\transp \mX_2 - \mX_2^\transp \mX_1 \inv{\mX_1^\transp \mX_1} \mX_1^\transp \mX_2} \\
    &= \inv{\mX_2^\transp
    \bPi_{\mathrm{Null}(\mX_1^\transp)} \mX_2},
\end{align}
Where $\bPi_{\mathrm{Null}(\mX_1^\transp)} \defeq \mI_n - \left(\mX_1^\transp\right)^\dagger \mX_1^\transp$ denotes the projection onto the column space of $\mX_1$. This gives
\begin{align}
    \mY_2 &= \mC \mX_1^\transp + \mD \mX_2^\transp \\
    &= \mD \mX_2^\transp \left( \mI_n - \mX_1 \inv{\mX_1^\transp \mX_1} \mX_1^\transp \right) \\
    &= \mD \mX_2^\transp \bPi_{\mathrm{Null}(\mX_1^\transp)}
    \left(
    \mX_2 \mX_2^\dagger + \bPi_{\mathrm{Null}(\mX_2^\transp)}
    \right) \\
    \label{eq:lemma:subselect-smaller:y2}
    &= \mX_2^\dagger + \mD \mX_2^\transp \bPi_{\mathrm{Null}(\mX_1^\transp)} \bPi_{\mathrm{Null}(\mX_2^\transp)}.
\end{align}
Let $\mU$, $\mU_*$, and $\mV$ be the matrices containing the left singular vectors of $\mX_2$, $\bPi_{\mathrm{Null}(\mX_2^\transp)}$, and $\bPi_{\mathrm{Null}(\mX_1^\transp)}$, respectively. Because the rows of $\mX$ are independently drawn from a spherical Gaussian distribution, $\mV$ has a uniform distribution over orthogonal matrices in $\reals^{n \times |S^c|}$. As such, $\E_\mV \left[ \mV^\transp \mU_* | \mV^\transp \mU \right] = \bm{0}$. Then
\begin{align}
    \E_{\mX_1} \left[
        \mD \mX_2^\transp \bPi_{\mathrm{Null}(\mX_1^\transp)} \bPi_{\mathrm{Null}(\mX_2^\transp)}
    \right] 
    &= \E_\mV \left[
    \inv{
    \mX_2^\transp \mV \mV^\transp \mX_2
    }
    \mX_2^\transp \mV \mV^\transp \mU_* \mU_*^\transp
    \right] \\
    &= \E_\mV \left[
    \inv{
    \mX_2^\transp \mV \mV^\transp \mX_2
    }
    \mX_2^\transp \mV \E_\mV \left[ \mV^\transp \mU_* \Big| \mV^\transp \mU \right] \mU_*^\transp
    \right] \\
    &= \bm{0},
\end{align}
which combined with \eqref{eq:lemma:subselect-smaller:y2} yields the first claim.

For the second claim, let $\mV_*$ denote the left singular vectors of $\mX_1$, and observe that \linebreak $\E_\mV \left[ \mV^\transp \mU_* | \mV^\transp \mU, \mV_* \right] = \bm{0}$. Then using similar arguments,
\begin{align}
    &\E_{\mX_1} \left[
    \mX_1^\transp f(\mX_2) \mS^\transp \mX^\dagger
    \right] \nonumber \\
    &= \E_{\mX_1} \left[
    \mX_1^\transp f(\mX_2) \mS^\transp \left(\mX_2^\dagger + \mD \mX_2^\transp 
    \bPi_{\mathrm{Null}(\mX_1^\transp)} \bPi_{\mathrm{Null}(\mX_2^\transp)}
    \right)
    \right] \\
    &= \E_{\mX_1} \left[
    \mX_1^\transp f(\mX_2) \mS^\transp \left(\mX_2^\dagger +
    \inv{
    \mX_2^\transp \mV \mV^\transp \mX_2
    }
    \mX_2^\transp \mV \E_\mV \left[ \mV^\transp \mU_* \Big| \mV^\transp \mU, \mV_* \right] \mU_*^\transp
    \right)
    \right] \\
    &= \E_{\mX_1} \left[
    \mX_1^\transp f(\mX_2) \mS^\transp \mX_2^\dagger
    \right] \\
    &= \bm{0}.
\end{align}

\subsection{Proof of Lemma~\ref{lemma:subselect-larger}}

Define $\bPi_{\mathrm{Null}(\mX^\transp)} \defeq \mI_n - \left(\mX^\transp\right)^\dagger \mX^\transp$, the projection operator onto the null space of $\mX^\transp$. Then for the first claim,
\begin{align}
    &\E_{T_1, T_2} \left[
    \left( \mT_1^\transp \mX \right)^\dagger \mT_1^\transp 
    \left( \left( \mT_2^\transp \mX \right)^\dagger \mT_2^\transp \right)^\transp
    \right] \nonumber \\
    &= 
    \E_{T_1, T_2} \left[
    \inv{\mX^\transp \mT_1 \mT_1^\transp \mX}
    \mX^\transp \mT_1 \mT_1^\transp
    \mT_2 \mT_2^\transp \mX
    \inv{\mX^\transp \mT_2 \mT_2^\transp \mX}
    \right] \\
    &= \E_{T_1, T_2} \left[
    \inv{\mX^\transp \mT_1 \mT_1^\transp \mX}
    \mX^\transp \mT_1 \mT_1^\transp
    \left( 
    \mX \left(\mX \mX^\transp\right)^\dagger \mX^\transp
    + \bPi_{\mathrm{Null}(\mX^\transp)}
    \right)
    \mT_2 \mT_2^\transp \mX
    \inv{\mX^\transp \mT_2 \mT_2^\transp \mX}
    \right] \\
    &= 
    \left(\mX \mX^\transp\right)^\dagger + 
    \E_{T_1, T_2} \left[
    \inv{\mX^\transp \mT_1 \mT_1^\transp \mX}
    \mX^\transp \mT_1 \mT_1^\transp
    \bPi_{\mathrm{Null}(\mX^\transp)}
    \mT_2 \mT_2^\transp \mX
    \inv{\mX^\transp \mT_2 \mT_2^\transp \mX}
    \right] \\
    \label{eq:lemma:subselect-larger:condition-on-set-size}
    &=
    \left(\mX \mX^\transp\right)^\dagger + 
    \frac{|T_1||T_2|}{n^2} \inv{\mX^\transp \mT_1 \mT_1^\transp \mX}
    \mX^\transp 
    \bPi_{\mathrm{Null}(\mX^\transp)}
    \mX
    \inv{\mX^\transp \mT_2 \mT_2^\transp \mX}
    \\
    \label{eq:lemma:subselect-larger:null-space}
    &= 
    \left(\mX^\transp \mX \right)^\dagger.
\end{align}
The equality~\eqref{eq:lemma:subselect-larger:condition-on-set-size} follows due the fact that, because of the distributional assumption on the rows of $\mX$, $\mX^\transp \mT_j \mT_j^\transp \mX$ and $\mT_j \mT_j^\transp$ are conditionally independent given $|T_j|$. The equality~\eqref{eq:lemma:subselect-larger:null-space} follows because $\mX^\transp \bPi_{\mathrm{Null}(\mX^\transp)} = \bf{0}$.

For the second claim,
\begin{align}
    &\E_{T_1, T_2} \left[
    \left( \left( \mT_1^\transp \mX \right)^\dagger \mT_1^\transp \right)^\transp
    \mA
    \left( \mT_2^\transp \mY \right)^\dagger \mT_2^\transp 
    \right] \nonumber \\
    &= 
    \E_{T_1, T_2} \left[
    \mT_1 \mT_1^\transp \mX
    \inv{\mX^\transp \mT_1 \mT_1^\transp \mX}
    \mA
    \inv{\mY^\transp \mT_2 \mT_2^\transp \mY}
    \mY^\transp \mT_2 \mT_2^\transp
    \right] \\
    &= 
    \E_{T_1, T_2} \left[
    \left(
    \left(\mX^\transp\right)^\dagger \mX^\transp
    + 
    \bPi_{\mathrm{Null}(\mX^\transp)}
    \right)
    \mT_1 \mT_1^\transp \mX
    \inv{\mX^\transp \mT_1 \mT_1^\transp \mX}
    \mA
    \inv{\mY\transp \mT_2 \mT_2^\transp \mY}
    \mY^\transp \mT_2 \mT_2^\transp
    \right] \\
    &= 
    \label{eq:lemma:subselect-larger:condition-and-null-1}
    \E_{\Pi(T_2)} \left[
    \left(\mX^\dagger \right)^\transp
    \mA
    \inv{\mY^\transp \mT_2 \mT_2^\transp \mY}
    \mY^\transp \mT_2 \mT_2^\transp
    \right] \\
    &= 
    \E_{\Pi(T_2)} \left[
    \left(\mX^\dagger \right)^\transp
    \mA
    \inv{\mY^\transp \mT_2 \mT_2^\transp \mY}
    \mY^\transp \mT_2 \mT_2^\transp
    \left(
    \mY \mY^\dagger
    + 
    \bPi_{\mathrm{Null}(\mY^\transp)}
    \right)
    \right] \\
    &= 
    \label{eq:lemma:subselect-larger:condition-and-null-2}
    \left(\mX^\dagger \right)^\transp
    \mA
    \mY^\dagger,
\end{align}
where the equations \eqref{eq:lemma:subselect-larger:condition-and-null-1} and \eqref{eq:lemma:subselect-larger:condition-and-null-2} follow by similar arguments to those used to show the first claim.

\end{document}